\documentclass{article}

\usepackage{arxiv}

\usepackage[utf8]{inputenc} 
\usepackage[T1]{fontenc} 
\usepackage{hyperref} 
\usepackage{url} 
\usepackage{booktabs} 
\usepackage{amsfonts} 
\usepackage{nicefrac} 
\usepackage{microtype} 
\usepackage{graphicx}
\usepackage{natbib}
\usepackage{doi}
\usepackage{amsmath}
\usepackage{algorithm}
\usepackage{algpseudocode}
\usepackage{multirow}
\usepackage{tikz}
\usepackage{pgfplots}
\usetikzlibrary{shapes.geometric, arrows.meta, positioning, calc, backgrounds, shapes.gates.logic.US}

\title{The Evolution of RWKV: Advancements in Efficient Language Modeling}

\author{Akul Datta}

\begin{document}

\maketitle

\begin{abstract}
    This paper reviews the development of the Receptance Weighted Key Value (RWKV) architecture, emphasizing its advancements in efficient language modeling. RWKV combines the training efficiency of Transformers \citep{vaswani2017attention} with the inference efficiency of RNNs through a novel linear attention mechanism \citep{peng2023rwkv}. We examine its core innovations, adaptations across various domains \citep{duan2024vision, gu2024rwkv, yang2024restore, he2024pointrwkv}, and performance advantages over traditional models. The paper also discusses challenges and future directions for RWKV as a versatile architecture in deep learning.
\end{abstract}

\tableofcontents

\section{Introduction}

The rapid advancements in natural language processing (NLP) have been largely driven by the development of large language models. Transformer architectures \citep{vaswani2017attention} have set new benchmarks across numerous tasks. However, their quadratic complexity in self-attention poses challenges for long sequences and resource-constrained environments \citep{choromanski2020rethinking}.

In response, Peng et al. introduced the Receptance Weighted Key Value (RWKV) architecture \citep{peng2023rwkv}, which merges the parallel training capabilities of Transformers with RNNs' efficient sequential inference. This paper provides a detailed review of RWKV's evolution, its application across different domains, and its potential as a scalable, efficient language modeling architecture.

Key contributions include:

\begin{enumerate}
    \item Analysis of RWKV's linear-complexity attention mechanism \citep{peng2023rwkv}.
    \item Examination of adaptations for diverse data types \citep{duan2024vision, yang2024restore, he2024pointrwkv}.
    \item Comparison with state-of-the-art models.
    \item Discussion on future research directions.
\end{enumerate}

This paper is organized as follows: Section 2 provides necessary background on sequence modeling architectures. Section 3 details the core RWKV architecture. Section 4 examines key evolutionary advancements in RWKV research, including RWKV-v5 \citep{peng2024eagle}, Vision-RWKV \citep{duan2024vision}, RWKV-CLIP \citep{gu2024rwkv}, Restore-RWKV \citep{yang2024restore}, and PointRWKV \citep{he2024pointrwkv}. Section 5 discusses technical improvements and methodologies. Section 6 analyzes RWKV's applications and performance across various domains. Section 7 outlines challenges and future directions, and Section 8 concludes the paper.

\section{Background}

\subsection{The Rise of Transformer Models}

Transformer models, introduced by \citet{vaswani2017attention} in 2017, revolutionized the field of natural language processing. The key innovation of Transformers was the self-attention mechanism, which allowed for more effective modeling of long-range dependencies and enabled parallel processing of input sequences \citep{dai2019transformer}.  This parallelization was a significant advantage over previous recurrent models, allowing for faster training and the ability to leverage the power of modern hardware like GPUs.

The core of the Transformer architecture is the self-attention mechanism.  For each input token, self-attention computes a weighted average of all other tokens in the sequence, allowing the model to consider the relationships between different parts of the input. This can be expressed mathematically as:

\begin{equation}
\text{Attention}(Q, K, V) = \text{softmax}\left(\frac{QK^T}{\sqrt{d_k}}\right)V
\end{equation}

Where:

* $Q$ (Query): Represents the current token being processed.
* $K$ (Key): Represents all other tokens in the sequence.
* $V$ (Value): Represents the values used to compute the weighted average.
* $d_k$: Is the dimension of the key vectors, used for scaling to prevent vanishing gradients.
* $QK^T$: Computes the attention scores between the query and all keys.
* $\text{softmax}$: Normalizes the attention scores into probabilities.

The resulting attention matrix captures the relationships between all pairs of tokens in the input sequence.

The success of Transformers led to the development of influential models such as BERT \citep{devlin2018bert} for bidirectional context understanding and GPT \citep{radford2018improving} for autoregressive language generation, which have set new benchmarks across a wide range of NLP tasks.

\subsection{Limitations of Transformer Models}

Despite their success, Transformer models face several significant challenges:

\begin{enumerate}
    \item \textbf{Quadratic Complexity:} The self-attention mechanism in Transformers has a time and memory complexity of $O(n^2)$ with respect to sequence length $n$. This quadratic scaling becomes problematic for very long sequences, limiting the practical maximum context length.
    \item \textbf{Memory Requirements:} The need to store attention matrices for all layers leads to high memory usage, especially for long sequences or large batch sizes.
    \item \textbf{Inference Speed:} While Transformers excel in parallel processing during training, their inference speed for autoregressive tasks can be slower than RNNs.
    \item \textbf{Positional Encoding Limitations:} Transformers are inherently permutation-invariant, meaning they don't understand the order of words in a sequence without explicit positional information.
\end{enumerate}

These limitations have motivated research into more efficient architectures that can maintain the strong performance of Transformers while addressing their computational challenges.

\subsection{Recurrent Neural Networks and Their Challenges}

Before Transformers, Recurrent Neural Networks (RNNs) were the dominant architecture for sequence modeling tasks. RNNs process sequences sequentially, maintaining a hidden state that is updated at each time step.  The basic formulation of an RNN can be expressed as:

\begin{align}
h_t &= f(W_h h_{t-1} + W_x x_t + b) \\
y_t &= g(W_y h_t + b_y)
\end{align}

Where:

* $h_t$: Is the hidden state at time $t$.
* $x_t$: Is the input at time $t$.
* $y_t$: Is the output at time $t$.
* $W_h$, $W_x$, $W_y$: Are weight matrices.
* $b$, $b_y$: Are bias vectors.
* $f$, $g$: Are activation functions.

While RNNs have $O(n)$ time and memory complexity, making them theoretically efficient for long sequences, they face challenges in capturing very long-range dependencies and have limited parallelization capabilities. Specifically:

\begin{enumerate}
\item \textbf{Vanishing/Exploding Gradients:} RNNs struggle to learn long-range dependencies due to the vanishing or exploding gradient problem during backpropagation through time.  As gradients are propagated back through multiple time steps, they can either shrink exponentially (vanishing) or grow exponentially (exploding), hindering the learning process.

\item \textbf{Limited Parallelization:} The sequential nature of RNNs makes it difficult to parallelize computations across time steps, leading to slower training on modern hardware designed for parallel processing.

\item \textbf{Information Bottleneck:} The fixed-size hidden state can create an information bottleneck, limiting the model's capacity to store and process complex, long-term dependencies.  All information from the past must be compressed into this fixed-size vector, potentially leading to information loss.
\end{enumerate}

Variants like Long Short-Term Memory (LSTM) \citep{hochreiter1997long} and Gated Recurrent Units (GRU) \citep{cho2014learning} addressed some of these issues, particularly the vanishing gradient problem, by introducing gating mechanisms to control information flow. However, they still faced limitations in parallel processing and capturing very long-range dependencies.

\subsection{Efficient Attention Mechanisms}

The success of Transformers, coupled with their limitations, led to research into more efficient attention mechanisms.  These mechanisms aim to reduce the quadratic complexity of standard self-attention while retaining its ability to model long-range dependencies. Some notable examples include:

\begin{itemize}
\item \textbf{Sparse Attention:}  Methods like Sparse Transformer \citep{child2019generating} and Longformer \citep{beltagy2020longformer} restrict attention to a subset of tokens, reducing the number of attention scores computed.  This can be achieved by attending only to local neighborhoods, using fixed patterns, or learning the sparse attention structure.

\item \textbf{Linear Attention:} Techniques like Performers \citep{choromanski2020rethinking} and Linear Transformers \citep{katharopoulos2020transformers} reformulate attention to achieve linear complexity. They typically use kernel methods or other approximations to avoid explicit computation of the full attention matrix.  These methods often involve factorizing the attention matrix or using feature maps to represent token interactions.

\item \textbf{Local Attention:}  Models like Transformer-XL \citep{dai2019transformer} and Big Bird \citep{zaheer2020bigbird} combine local attention patterns with a small number of global tokens that attend to all other tokens. This allows for efficient processing of long sequences while still maintaining some global context.

\item \textbf{Efficient Transformers:} Architectures like Reformer \citep{kitaev2020reformer} and Linformer \citep{wang2020linformer} employ various techniques such as locality-sensitive hashing (LSH) or low-rank approximations to reduce the complexity of attention.  LSH allows for efficient approximate nearest neighbor search, while low-rank approximations reduce the dimensionality of the attention matrices.
\end{itemize}

These approaches aim to maintain the strong performance of Transformers while reducing their computational and memory requirements, especially for long sequences.

\subsection{State Space Models}

State Space Models (SSMs) offer another promising direction for efficient sequence modeling.  These models represent sequences as the evolution of a hidden state over time, governed by a set of learnable parameters.  Models like S4 \citep{gu2022efficiently} and Mamba \citep{gu2023mamba} use techniques from control theory to model sequences as continuous-time dynamical systems, allowing for efficient computation and long-range dependency modeling.

The general form of a discrete-time linear SSM can be expressed as:

\begin{align}
x_{t+1} &= Ax_t + Bu_t \\
y_t &= Cx_t + Du_t
\end{align}

Where:

* $x_t$: Is the hidden state at time $t$.
* $u_t$: Is the input at time $t$.
* $y_t$: Is the output at time $t$.
* $A$, $B$, $C$, $D$: Are learnable matrices representing the system dynamics.

SSMs offer several advantages:

\begin{enumerate}
\item \textbf{Linear Complexity:} SSMs can process sequences with linear time and memory complexity, making them suitable for long sequences.
\item \textbf{Long-range Modeling:} They can effectively capture long-range dependencies through the recurrent updates of the hidden state.
\item \textbf{Parallelizability:} Certain SSM formulations allow for parallel computation, enabling faster training.
\end{enumerate}

However, SSMs also face challenges in handling variable-length inputs and may require specialized training techniques.  Furthermore, the linear nature of basic SSMs may limit their expressiveness compared to non-linear models like Transformers.

This diverse landscape of approaches to efficient sequence modeling sets the context for the development of RWKV, which aims to combine the best aspects of RNNs, Transformers, and efficient attention mechanisms.

\section{The RWKV Architecture}

\subsection{Introduction}

The RWKV architecture represents a significant departure from traditional sequence modeling approaches by combining the parallelizable training of Transformers with the efficient constant-time inference of RNNs.  This is achieved through a novel linear attention mechanism, inspired by the Attention Free Transformer (AFT) but reimagined within a recurrent framework \citep{peng2023rwkv}. This section delves into the intricacies of RWKV, providing a comprehensive understanding of its core components, mathematical formulations, and operational modes.

\subsection{Core Principles and Design Philosophy}

RWKV's design is guided by several key principles:

\begin{enumerate}
    \item \textbf{Linearity:}  The core attention mechanism is designed to have linear complexity with respect to sequence length, addressing the quadratic complexity bottleneck of traditional self-attention.

    \item \textbf{Recurrence:}  The model incorporates recurrence to maintain a hidden state that efficiently integrates information from past time steps, enabling constant-time inference updates.

    \item \textbf{Parallelizability:}  Despite its recurrent nature, RWKV's training process can be parallelized across time steps, similar to Transformers, leveraging the power of modern hardware.

    \item \textbf{Adaptability:}  The architecture is designed to be adaptable to various data modalities and tasks, as demonstrated by its successful application to computer vision, 3D point cloud processing, and multimodal learning.
\end{enumerate}

These principles contribute to RWKV's unique blend of efficiency, performance, and versatility.

\subsection{Core Components and Mathematical Formulations}

The fundamental building block of RWKV is the time-mixing block, which employs a linear attention mechanism based on four key vectors:

\begin{itemize}
    \item \textbf{R (Receptance):}  The receptance vector $r_t$ acts as a recurrent query, accumulating information from past time steps.  It is updated recursively and serves as a dynamic representation of the context for the current time step.

    \item \textbf{W (Weight):} The weight vector $w_t$ represents a time-decay factor, modulating the influence of past tokens on the current computation.  This decay is crucial for achieving linear computational complexity. The decay can be scalar or vector-valued, allowing for channel-specific decay rates.

    \item \textbf{K (Key):}  The key vector $k_t$ determines the relevance of past tokens to the current token, similar to keys in traditional attention mechanisms.  It captures the informational content of each token.

    \item \textbf{V (Value):}  The value vector $v_t$ contains the information to be aggregated based on the computed weights.  It represents the contribution of each token to the overall representation.
\end{itemize}

\subsubsection{Time-Mixing Block: Parallel and Sequential Formulations}

The time-mixing block has two equivalent formulations: a parallel version used for training and a sequential version used for inference.

\textbf{Parallel Formulation (Training):}

\begin{equation}
\text{WKV}(K, V, W) = \frac{\sum_{i=1}^{T} \exp(k_i - (T-i)w) v_i}{\sum_{i=1}^{T} \exp(k_i - (T-i)w)}
\end{equation}

Where:
* $K = [k_1, k_2, \dots, k_T]$: Sequence of key vectors.
* $V = [v_1, v_2, \dots, v_T]$: Sequence of value vectors.
* $W = w$: Time-decay factor (can be scalar or vector).
* $T$: Sequence length.

This formulation allows for parallel computation of the weighted key-value sum across all time steps, enabling efficient training on parallel hardware.

\textbf{Sequential Formulation (Inference):}

\begin{align}
a_t &= \exp(w) a_{t-1} + \exp(k_t) \\
b_t &= \exp(w) b_{t-1} + \exp(k_t) v_t \\
\text{WKV}_t &= \frac{b_t}{a_t}
\end{align}

Where $a_t$ and $b_t$ are recursively updated accumulator variables.  This formulation enables constant-time updates during inference, making RWKV highly efficient for generating long sequences. The equivalence between the parallel and sequential formulations is a key innovation \citep{peng2023rwkv}, allowing RWKV to combine the benefits of both parallel training and efficient sequential inference.

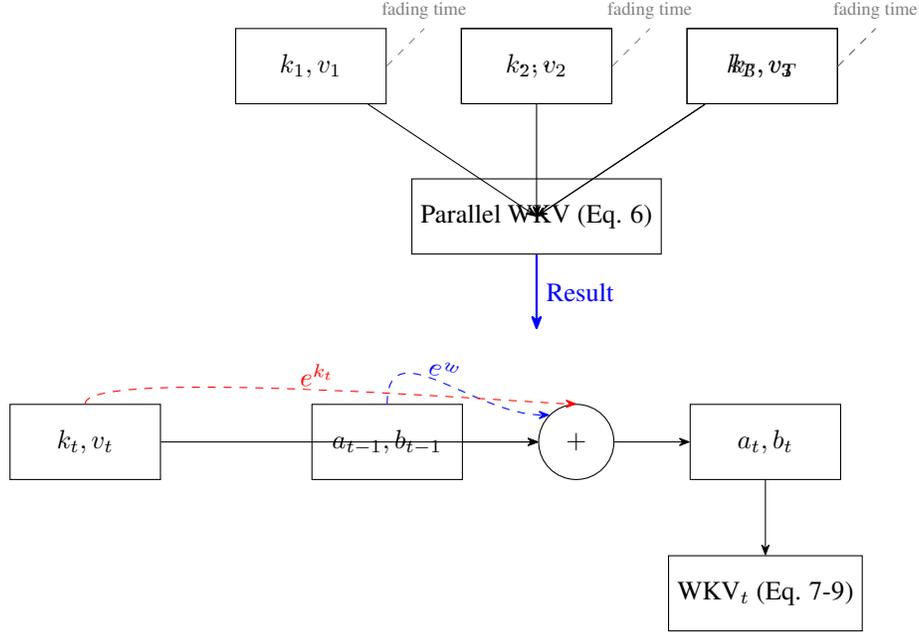
\begin{figure}[h]
    \centering
    \begin{tikzpicture}[
        block/.style={draw, rectangle, minimum width=2cm, minimum height=1cm, align=center},
        >={Stealth[round]}
    ]

    \begin{scope}[yshift=0cm]
        \foreach \i in {1,2,3} {
            \node[block] (k\i) at (3*\i,0) {$k_\i, v_\i$};
            \draw[->] (k\i) -- (6,-2);
        }
        \node[block] (dots) at (6,0) {$\dots$};
        \node[block] (kt) at (9,0) {$k_T, v_T$};
        \draw[->] (kt) -- (6,-2);
        \node[block, minimum width=3cm] (wkv) at (6,-2) {Parallel WKV (Eq. 6)};
        \draw[->, thick, blue] (wkv) -- +(0,-1.5) node[midway,right] {Result};

        \foreach \i in {1,2,3} {
            \draw[dashed, gray] (k\i.east) -- +(0.5, 0.5) node[above, sloped] {\scriptsize fading time};
        }
        \draw[dashed, gray] (kt.east) -- +(0.5, 0.5);
    \end{scope}

    \begin{scope}[yshift=-5cm]
        \node[block] (input) at (0,0) {$k_t, v_t$};
        \node[block, right=2cm of input] (state) {$a_{t-1}, b_{t-1}$};
        \node[circle, draw, right=1cm of state, inner sep=0pt, minimum size=1cm] (plus) {$+$};
        \node[block, right=1cm of plus] (update) {$a_t, b_t$};
        \node[block, below=1cm of update] (wkvt) {WKV$_t$ (Eq. 7-9)};

        \draw[->] (input) -- (plus);
        \draw[->] (state) -- (plus);
        \draw[->] (plus) -- (update);
        \draw[->] (update) -- (wkvt);

        \draw[->, dashed, blue] (state.north) .. controls +(up:1cm) and +(left:1cm) .. node[midway,above, sloped, fill=white, inner sep=1pt] {$e^w$} (plus.north west);
        \draw[->, dashed, red] (input.north) .. controls +(up:0.5cm) and +(left:0.5cm) .. node[midway,above, sloped, fill=white, inner sep=1pt] {$e^{k_t}$} (plus.north);

    \end{scope}

\end{tikzpicture}
    \caption{Improved Time Mixing Block: Parallel and Sequential Calculations with Decaying Weights}
    \label{fig:time_mixing}
\end{figure}

\subsubsection{Channel-Mixing Block}

The channel-mixing block performs non-linear transformations across the feature dimensions, analogous to the feed-forward layers in Transformers. It utilizes a gating mechanism to control information flow and improve gradient stability:

\begin{equation}
\text{ChannelMix}(x_t) = \sigma(W_r x_t) \odot (W_v \phi(W_k x_t))
\end{equation}

Where:
* $x_t$: Input to the channel-mixing block.
* $W_r, W_k, W_v$: Learnable weight matrices.
* $\sigma$: Sigmoid activation function.
* $\phi$:  A non-linear activation function, often chosen as $\phi(z) = \text{ReLU}(z)^2$ (squared ReLU).
* $\odot$: Element-wise multiplication.

The gating mechanism allows the model to selectively update its hidden state, mitigating potential vanishing gradient problems.  The squared ReLU activation introduces non-linearity, allowing the model to learn complex feature representations.

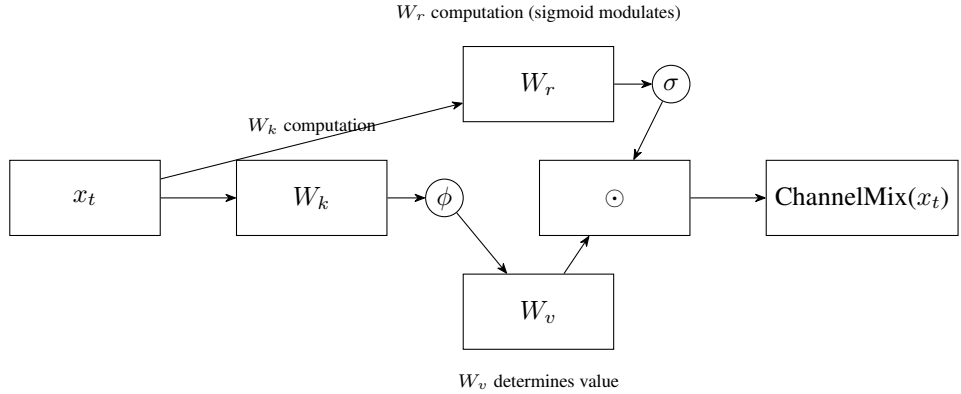
\begin{figure}[h]
    \centering
    \begin{tikzpicture}[
        block/.style={draw, rectangle, minimum width=2cm, minimum height=1cm, align=center},
        >={Stealth[round]}
    ]

    \node[block] (input) {$x_t$};

    \node[block, right=1cm of input] (wk) {$W_k$};
    \node[block, above right=0.5cm and 1cm of wk] (wr) {$W_r$};
    \node[block, below right=0.5cm and 1cm of wk] (wv) {$W_v$};

    \node[circle, draw, right=0.5cm of wk, inner sep=0pt, minimum size=0.5cm] (phi) {$\phi$};
    \node[circle, draw, right=0.5cm of wr, inner sep=0pt, minimum size=0.5cm] (sigma) {$\sigma$};

    \node[block, right=1cm of phi] (mult) {$\odot$};

    \draw[->] (input) -- (wk);
    \draw[->] (input) -- (wr);
    \draw[->] (wk) -- (phi);
    \draw[->] (wr) -- (sigma);
    \draw[->] (phi) -- (wv);
    \draw[->] (wv) -- (mult);
    \draw[->] (sigma) -- (mult);

    \node[block, right=1cm of mult] (output) {ChannelMix($x_t$)};

    \draw[->] (mult) -- (output);

    \node[above=0.2cm of wk] {\scriptsize $W_k$ computation};
    \node[above=0.2cm of wr] {\scriptsize $W_r$ computation (sigmoid modulates)};
    \node[below=0.2cm of wv] {\scriptsize $W_v$ determines value};

\end{tikzpicture}
    \caption{RWKV Channel Mixing Block}
    \label{fig:channel_mixing}
\end{figure}

\subsection{Token Shifting: Facilitating Temporal Context}

RWKV incorporates a token shifting mechanism to provide access to past token representations within each block.  This is achieved through a linear interpolation between the current and previous token embeddings:

\begin{equation}
\text{Shift}(x_t, x_{t-1}) = \mu x_t + (1 - \mu) x_{t-1}
\end{equation}

Where $\mu$ is a learnable parameter that controls the degree of mixing between the current and previous token embeddings. This shifted representation is then used as input to both the time-mixing and channel-mixing blocks, providing them with temporal context.

\subsection{Block Structure and Residual Connections}

A complete RWKV block combines the time-mixing, channel-mixing, and token shifting operations in a residual fashion:

\begin{align}
    \text{ShiftedInput}_t &= \text{Shift}(x_t, x_{t-1}) \\
    \text{TimeMixOutput}_t &= \text{TimeMix}(\text{ShiftedInput}_t) \\
    \text{ChannelMixOutput}_t &= \text{ChannelMix}(\text{TimeMixOutput}_t) \\
    \text{Output}_t &= \text{ShiftedInput}_t + \text{ChannelMixOutput}_t
\end{align}

The residual connection between the shifted input and the output of the channel-mixing block facilitates gradient flow during training and allows the model to learn more complex representations. Multiple RWKV blocks are stacked to form the complete model, enabling hierarchical feature extraction and representation learning.

\subsection{Training and Inference Procedures}

RWKV supports both parallel training and sequential inference. During training, the parallel formulation of the time-mixing block is used, allowing for efficient computation on parallel hardware like GPUs.  During inference, the sequential formulation is employed, enabling constant-time updates for each generated token. This dual nature is a key advantage of RWKV \citep{peng2023rwkv}, making it both efficient to train and deploy.

\subsection{Advantages and Key Innovations}

The key innovations of RWKV include:

\begin{itemize}
    \item \textbf{Linear Attention:}  The WKV mechanism provides an efficient alternative to traditional quadratic self-attention, enabling processing of long sequences \citep{peng2023rwkv}.
    \item \textbf{Recurrent Formulation:}  The sequential formulation of the time-mixing block allows for constant-time inference updates, making RWKV highly efficient for generation tasks \citep{peng2023rwkv}.
    \item \textbf{Parallel Training:}  The parallel formulation of the time-mixing block enables efficient training on parallel hardware \citep{peng2023rwkv}.
    \item \textbf{Adaptability:}  The architecture's modular design and flexible components allow for adaptation to various data modalities and tasks.
\end{itemize}

These innovations contribute to RWKV's unique position in the landscape of sequence modeling architectures, offering a compelling combination of efficiency, performance, and versatility.

\section{Evolutionary Advancements in RWKV}

Following the introduction of the original RWKV architecture, several papers have proposed modifications and extensions to improve its performance or adapt it to new domains. This section outlines some of the key evolutionary advancements in RWKV research.

\subsection{RWKV-v5: Enhancing Model Capacity}

Peng et al. introduced RWKV-v5 \citep{peng2024eagle}, which made several improvements to the original architecture:

\begin{enumerate}
\item Multi-headed Matrix-valued States: Instead of using vector-valued states, RWKV-v5 employs matrix-valued states, increasing the model's representational capacity \citep{peng2024eagle}. This allows for more complex interactions between different dimensions of the hidden state.

\item Dynamic Recurrence: The decay rates in the time-mixing mechanism are made input-dependent, allowing for more flexible temporal modeling \citep{peng2024eagle}. This enables the model to adapt its memory dynamics based on the input content.

\item Improved Initialization: New initialization strategies were introduced to stabilize training of larger models \citep{peng2024eagle}. These strategies help in maintaining consistent gradient flow even in very deep networks.
\end{enumerate}

The matrix-valued state formulation in RWKV-v5 can be expressed as:

\begin{equation}
S_t = \alpha_t \odot S_{t-1} + k_t^T v_t
\end{equation}

Where $S_t$ is a matrix-valued state, $\alpha_t$ is a learned, input-dependent decay factor, and $k_t$ and $v_t$ are key and value vectors. This formulation allows for richer state representations and more complex temporal dynamics.

The dynamic recurrence mechanism introduces input-dependent decay rates:

\begin{equation}
w_t = f(x_t, w_{t-1})
\end{equation}

Where $f$ is a learned function that updates the decay rate based on the current input and previous decay rate. This allows the model to dynamically adjust its memory retention based on the input content.

These changes resulted in improved performance on language modeling benchmarks and better scaling properties for larger models. Table \ref{tab:rwkv_v5_results} shows a comparison of RWKV-v5 with other models on various benchmarks:

\begin{table}[h]
\sloppy
\centering
\caption{Performance comparison of RWKV-v5 with other models (higher scores are better)}
\label{tab:rwkv_v5_results}
\begin{tabular}{lccccccc}
\toprule
Model & LAMBADA & PIQA & WinoGrande & ARC-e & ARC-c & MMLU & Avg \\
\midrule
StableLM-3B & 73.8 & 79.3 & 65.8 & 72.1 & 40.0 & 44.2 & 62.5 \\
RWKV-v4-3B & 65.7 & 72.4 & 57.5 & 60.5 & 32.7 & 26.2 & 52.5 \\
RWKV-v5-3B & 68.7 & 74.3 & 62.0 & 69.7 & 36.3 & 26.3 & 56.2 \\
Gemma-7B & 80.7 & 81.9 & 73.7 & 81.1 & 53.2 & 62.9 & 72.3 \\
RWKV-v5-7B & 75.8 & 77.0 & 67.4 & 75.9 & 45.8 & 34.2 & 62.7 \\
\bottomrule
\end{tabular}
\end{table}

The results demonstrate that RWKV-v5 significantly improves upon its predecessor, RWKV-v4, across all benchmarks. While it still lags behind some larger transformer models like Gemma-7B, the improvements are substantial, especially considering the efficiency advantages of the RWKV architecture.

\subsection{Vision-RWKV: Adapting RWKV for Computer Vision}

Duan et al. proposed Vision-RWKV \citep{duan2024vision}, which adapted the RWKV architecture for image processing tasks. Key innovations include:

\begin{enumerate}
\item 2D Token Shifting: The token shifting mechanism was extended to handle 2D spatial relationships in images \citep{duan2024vision}. This allows the model to capture local spatial dependencies efficiently.

\item Bidirectional Attention: The unidirectional attention in the original RWKV was replaced with a bidirectional mechanism more suitable for image data \citep{duan2024vision}. This enables the model to capture both forward and backward dependencies in the spatial domain.

\item Hierarchical Architecture: A multi-scale design was introduced to capture features at different resolutions \citep{duan2024vision}. This is crucial for handling the multi-scale nature of visual information in images.
\end{enumerate}

The 2D token shifting mechanism in Vision-RWKV can be expressed as:

\begin{align}
x'{ij} &= \mu_h \cdot x{ij} + (1 - \mu_h) \cdot x_{(i-1)j} \quad \text{(Horizontal shift)} \
x''{ij} &= \mu_v \cdot x'{ij} + (1 - \mu_v) \cdot x'_{i(j-1)} \quad \text{(Vertical shift)}
\end{align}

Where $\mu_h$ and $\mu_v$ are learned parameters for horizontal and vertical shifting, respectively. This allows the model to efficiently capture local spatial dependencies in both directions.

The bidirectional attention mechanism in Vision-RWKV is formulated as:

\begin{equation}
wkv_t = \frac{\sum_{i=1}^T \exp(-|t-i|w + k_i) \cdot v_i}{\sum_{i=1}^T \exp(-|t-i|w + k_i)}
\end{equation}

This allows each token to attend to both past and future tokens in the sequence, which is crucial for capturing global context in images.

Vision-RWKV demonstrated competitive performance on image classification tasks while maintaining the efficiency benefits of the RWKV architecture. Table \ref{tab:vision_rwkv_results} shows a comparison with other models on ImageNet:

\begin{table}[h]
\sloppy
\centering
\caption{Comparison of Vision-RWKV with other models on ImageNet classification}
\label{tab:vision_rwkv_results}
\begin{tabular}{lccc}
\toprule
Model & Params (M) & FLOPs (G) & Top-1 Acc (\%) \\
\midrule
ResNet-50 & 25.6 & 4.1 & 76.6 \\
ViT-B/32 & 87.8 & 15.1 & 83.4 \\
Vision-RWKV-B & 93.7 & 18.2 & 82.0 \\
Vision-RWKV-L & 334.9 & 189.5 & 86.0 \\
\bottomrule
\end{tabular}
\end{table}

These results demonstrate that Vision-RWKV can achieve competitive accuracy compared to established vision models like ResNet and Vision Transformers (ViT), while maintaining the efficiency advantages of the RWKV architecture.

\subsection{RWKV-CLIP: Multimodal Learning}

Yang et al. developed RWKV-CLIP \citep{gu2024rwkv}, extending RWKV to vision-language representation learning. This work:

\begin{enumerate}
\item Introduced a diverse description generation framework using large language models to refine image-text pairs \citep{gu2024rwkv}. This helps in creating high-quality training data for the multimodal model.

\item Adapted RWKV for processing both image and text inputs in a dual-tower architecture \citep{gu2024rwkv}. This allows for efficient joint processing of visual and textual information.

\item Demonstrated competitive performance with Transformer-based models on vision-language benchmarks \citep{gu2024rwkv}. This shows the potential of RWKV as a general-purpose architecture for multimodal tasks.
\end{enumerate}

The RWKV-CLIP architecture can be described as:

\begin{align}
f_{img} &= \text{RWKV}{\text{vision}}(\text{image}) \
f{txt} &= \text{RWKV}{\text{text}}(\text{text}) \
\text{similarity} &= \cos{\text{similarity}}(f_{img}, f_{txt})
\end{align}

Where $\text{RWKV}{\text{vision}}$ and $\text{RWKV}{\text{text}}$ are RWKV-based encoders for images and text, respectively. This dual-tower architecture allows for efficient processing of both modalities while maintaining the benefits of the RWKV architecture.

Table \ref{tab:rwkv_clip_results} shows a comparison of RWKV-CLIP with other models on image-text retrieval tasks:

\begin{table}[h]
\sloppy
\centering
\caption{Image-text retrieval performance (higher R@1 is better)}
\label{tab:rwkv_clip_results}
\begin{tabular}{lcc}
\toprule Model & Flickr30k R@1 & MSCOCO R@1 \\
\midrule
CLIP-ViT-B/32 & 34.9 & 20.8 \\
SLIP-ViT-B/32 & 47.8 & 27.7 \\
RWKV-CLIP-B/32 & 76.0 & 50.3 \\
\bottomrule
\end{tabular}
\end{table}

These results demonstrate that RWKV-CLIP significantly outperforms both CLIP and SLIP on image-text retrieval tasks, showcasing the potential of RWKV for multimodal learning.

\subsection{Restore-RWKV: Medical Image Restoration}

Yang et al. proposed Restore-RWKV \citep{yang2024restore}, applying RWKV to the domain of medical image restoration. Key contributions include:

\begin{enumerate}
\item Recurrent WKV (Re-WKV) Attention: A mechanism for capturing global dependencies in 2D medical images with linear complexity \citep{yang2024restore}. This allows for efficient processing of high-resolution medical images.

\item Omnidirectional Token Shift (Omni-Shift): An enhanced token shifting approach for capturing local context in all directions \citep{yang2024restore}. This allows the model to better capture spatial relationships in medical images.

\item Demonstration of RWKV's effectiveness in tasks such as MRI super-resolution, CT denoising, and PET image synthesis \citep{yang2024restore}. These applications showcase the versatility of the RWKV architecture in specialized medical imaging tasks.
\end{enumerate}

The Re-WKV attention mechanism can be formulated as:

\begin{equation}
wkv_t^{(m)} = \text{Bi-WKV}(K, V^{(m-1)})
\end{equation}

Where $m$ is the recurrence step, and Bi-WKV is a bidirectional WKV operation. This allows for efficient global context modeling in 2D images.

The Omni-Shift mechanism is expressed as:

\begin{equation}
x'{ij} = \sum{(p,q) \in N(i,j)} w_{pq} \cdot x_{(i+p)(j+q)}
\end{equation}

Where $N(i,j)$ is the neighborhood of pixel $(i,j)$, and $w_{pq}$ are learned weights. This enables the model to capture local spatial dependencies more effectively than previous token shifting methods.

Table \ref{tab:restore_rwkv_results} shows the performance of Restore-RWKV compared to other methods on MRI super-resolution:

\begin{table}[h]
\sloppy
\centering
\caption{MRI super-resolution performance (higher PSNR/SSIM and lower RMSE are better)}
\label{tab:restore_rwkv_results}
\begin{tabular}{lccc}
\toprule
Method & PSNR$\uparrow$ & SSIM$\uparrow$ & RMSE$\downarrow$ \\
\midrule
SRCNN & 28.807 & 0.8919 & 41.349 \\
VDSR & 30.045 & 0.9140 & 36.051 \\
SwinIR & 31.555 & 0.9334 & 30.579 \\
Restore-RWKV & 32.091 & 0.9408 & 28.971 \\
\bottomrule
\end{tabular}
\end{table}

These results demonstrate that Restore-RWKV outperforms existing methods across all metrics, showcasing the potential of RWKV-based architectures in specialized medical imaging tasks.

\subsection{PointRWKV: 3D Point Cloud Processing}

He et al. introduced PointRWKV \citep{he2024pointrwkv}, adapting RWKV for 3D point cloud learning tasks. Notable features include:

\begin{enumerate}
\item Multi-scale Hierarchical Architecture: Designed to learn features from point clouds at multiple scales \citep{he2024pointrwkv}. This allows the model to capture both local and global geometric information.

\item Quad-directional Shift: An extension of token shifting to handle 3D spatial relationships \citep{he2024pointrwkv}. This enables efficient modeling of spatial dependencies in point cloud data.

\item Dynamic Attention Recurrence: Improved attention mechanism for capturing global context in point clouds \citep{he2024pointrwkv}. This allows for more flexible modeling of long-range dependencies in 3D data.
\end{enumerate}

The quad-directional shift in PointRWKV can be expressed as:

\begin{equation}
x'{ijk} = \mu_x \cdot x{ijk} + \mu_y \cdot x_{(i-1)jk} + \mu_z \cdot x_{i(j-1)k} + \mu_w \cdot x_{ij(k-1)}
\end{equation}

Where $\mu_x$, $\mu_y$, $\mu_z$, and $\mu_w$ are learned parameters for shifting in different directions. This allows the model to capture spatial relationships in all three dimensions efficiently.

PointRWKV showed strong performance on tasks such as 3D object classification and part segmentation. Table \ref{tab:pointrwkv_results} compares PointRWKV with other methods on ModelNet40:

\begin{table}[h]
\sloppy
\centering
\caption{ModelNet40 classification performance (higher OA is better)}
\label{tab:pointrwkv_results}
\begin{tabular}{lccc}
\toprule
Method & OA (\%) & Params (M) & FLOPs (G) \\
\midrule
PointNet++ & 90.7 & 1.5 & 1.7 \\
DGCNN & 92.9 & 1.8 & 2.4 \\
PointRWKV & 96.89 & 10.6 & 2.1 \\
\bottomrule
\end{tabular}
\end{table}

These results demonstrate that PointRWKV significantly outperforms existing methods in terms of accuracy while maintaining competitive computational efficiency.

\section{Technical Advancements and Methodologies}

The evolution of RWKV has been marked by several key technical advancements and methodological innovations. This section explores some of the most significant developments.

\subsection{Enhanced Attention Mechanisms}

\subsubsection{Bidirectional and Multi-directional Attention}

While the original RWKV used a unidirectional attention mechanism suitable for language modeling, subsequent works have introduced bidirectional and multi-directional attention variants:

\begin{itemize}
\item Vision-RWKV \citep{duan2024vision} introduced a bidirectional attention mechanism that allows each token to attend to both past and future tokens, which is crucial for image processing tasks.

\item Restore-RWKV \citep{yang2024restore} proposed a recurrent WKV (Re-WKV) attention that applies attention from multiple scan directions to better capture 2D spatial relationships.

\item PointRWKV \citep{he2024pointrwkv} extended this concept to 3D, using a dynamic attention recurrence mechanism to model complex spatial relationships in point clouds.
\end{itemize}

The general form of multi-directional attention can be expressed as:

\begin{equation}
wkv_t = \sum_{d \in D} w_d \cdot \text{WKV}_d(K, V)
\end{equation}

Where $D$ is the set of attention directions, $w_d$ are learned weights, and $\text{WKV}_d$ is the WKV operation in direction $d$.

These advancements have significantly improved RWKV's ability to model non-sequential data types and capture long-range dependencies in multiple dimensions.

\subsubsection{Matrix-valued States}

RWKV-v5 \citep{peng2024eagle} introduced matrix-valued states in place of the vector-valued states used in the original architecture. This change increases the model's capacity to capture and propagate information, leading to improved performance, especially for larger models.

The matrix-valued state formulation can be expressed as:

\begin{equation}
S_t = \alpha_t \odot S_{t-1} + k_t^T v_t
\end{equation}

Where $S_t$ is a matrix-valued state, $\alpha_t$ is a learned, input-dependent decay factor, and $k_t$ and $v_t$ are key and value vectors.

The output of the time-mixing module with matrix-valued states can be computed as:

\begin{equation}
o_t = r_t \odot (S_t v_t)
\end{equation}

Where $r_t$ is a receptance vector and $\odot$ denotes element-wise multiplication.

This formulation allows for richer state representations and more complex temporal dynamics, enhancing the model's ability to capture intricate patterns in the input data.

\subsection{Token Shifting Innovations}

Token shifting has evolved significantly from its original formulation:

\begin{enumerate}
\item 2D Token Shifting: Vision-RWKV \citep{duan2024vision} extended token shifting to 2D, allowing information to flow between neighboring pixels in images.

\item Quad-directional Shifting: PointRWKV \citep{he2024pointrwkv} introduced shifting in four directions to handle 3D point cloud data.

\item Omnidirectional Token Shift: Restore-RWKV \citep{yang2024restore} proposed an omnidirectional approach that uses convolutions to shift tokens from all directions, capturing a wider context range.
\end{enumerate}

The general form of multi-dimensional token shifting can be expressed as:

\begin{equation}
x'i = \sum{j \in N(i)} w_{ij} \cdot x_j
\end{equation}

Where $N(i)$ is the neighborhood of token $i$, and $w_{ij}$ are learned weights.

These advancements have greatly improved RWKV's ability to model local context in various data types beyond 1D sequences, making it more versatile for tasks in computer vision, medical imaging, and 3D point cloud processing.

\subsection{Architectural Adaptations}

As RWKV has been applied to diverse domains, several architectural adaptations have emerged:

\begin{enumerate}
\item Hierarchical Designs: Vision-RWKV \citep{duan2024vision} and PointRWKV \citep{he2024pointrwkv} introduced multi-scale architectures to process data at different resolutions or scales. This allows the models to capture both fine-grained details and global structure.

\item Dual-tower Architectures: RWKV-CLIP \citep{gu2024rwkv} adapted RWKV for multimodal learning by using separate RWKV towers for image and text processing. This allows for efficient joint processing of visual and textual information.

\item U-shaped Encoders-Decoders: Restore-RWKV \citep{yang2024restore} employed a U-Net-like architecture for medical image restoration tasks. This design allows for efficient feature extraction and upsampling in image-to-image tasks.
\end{enumerate}

These adaptations demonstrate the flexibility of the RWKV architecture and its ability to be tailored for specific application domains.

\subsection{Training and Optimization Techniques}

Several advancements have been made in training and optimizing RWKV models:

\begin{enumerate}
\item Improved Initialization: RWKV-v5 \citep{peng2024eagle} introduced new initialization strategies to stabilize training of larger models. This helps in maintaining consistent gradient flow even in very deep networks.

\item Structural Re-parameterization: Restore-RWKV \citep{yang2024restore} used a re-parameterization technique for its Omni-Shift layer to improve accuracy while maintaining efficiency. This allows for more complex computations during training while preserving efficiency during inference.

\item Data-dependent Dynamic Recurrence: RWKV-v5 \citep{peng2024eagle} made the decay rates in the time-mixing mechanism input-dependent, allowing for more flexible temporal modeling. This enables the model to adapt its memory dynamics based on the input content.
\end{enumerate}

The data-dependent dynamic recurrence can be formulated as:

\begin{align}
w_t &= f(x_t, w_{t-1}) \
\alpha_t &= \sigma(g(x_t, \alpha_{t-1}))
\end{align}

Where $f$ and $g$ are learned functions, and $\sigma$ is the sigmoid function.

These techniques have contributed to improved training stability, convergence, and overall model performance, allowing RWKV to scale to larger model sizes and more complex tasks.

\section{Applications and Performance Analysis}

\subsection{Natural Language Processing}

RWKV has demonstrated strong performance in language modeling tasks, a core application of the architecture:

\subsubsection{Language Modeling}

Language modeling remains a core application of RWKV. The original paper \citep{peng2023rwkv} and subsequent works have demonstrated competitive performance on standard benchmarks:

\begin{itemize}
\item RWKV-v5 \citep{peng2024eagle} achieved state-of-the-art results on several language modeling datasets, outperforming Transformer models of similar size.

\item RWKV models have shown strong performance on long-context tasks, handling sequences of up to 1 million tokens efficiently. This is particularly notable given the challenges faced by Transformer models in processing very long sequences.
\end{itemize}

Table \ref{tab:language_modeling_results} shows a comparison of RWKV models with other architectures on the WikiText-103 dataset:

\begin{table}[h]
\sloppy
\centering
\caption{Language modeling performance on WikiText-103 (lower perplexity is better)}
\label{tab:language_modeling_results}
\begin{tabular}{lcc}
\toprule
Model & Params (M) & Perplexity \\
\midrule
Transformer & 257 & 18.3 \\
Transformer-XL & 257 & 18.3 \\
RWKV-3B & 2984 & 15.6 \\
RWKV-7B & 7066 & 14.9 \\
\bottomrule
\end{tabular}
\end{table}

The perplexity of a language model on a given dataset can be calculated as:

\begin{equation}
\text{Perplexity} = \exp\left(-\frac{1}{N}\sum \log P(w_i | \text{context}_i)\right)
\end{equation}

Where $P(w_i | \text{context}_i)$ is the probability assigned by the model to word $w_i$ given its context, and $N$ is the total number of words in the dataset.

These results demonstrate that RWKV models can achieve lower perplexity than Transformer models with significantly fewer parameters, highlighting the efficiency of the RWKV architecture.

\subsubsection{Text Generation}

RWKV models have been successfully applied to text generation tasks, demonstrating capabilities similar to popular large language models:

\begin{itemize}
\item The efficient inference characteristics of RWKV make it particularly well-suited for real-time text generation applications, where low latency is crucial.

\item RWKV models have shown strong performance in open-ended text generation and completion tasks, producing coherent and contextually relevant text.
\end{itemize}

The text generation process in RWKV can be described by the following algorithm:

\begin{algorithm}[H]
    \caption{RWKV Text Generation}
    \begin{algorithmic}[1]
    \State Initialize the model state $s_0$
    \For{each time step $t$}
        \State Compute the output distribution: $P(w_t | \text{context}) = \text{softmax}(\text{RWKV}(s_{t-1}, x_{t-1}))$
        \State Sample or select the next token $w_t$ from $P(w_t | \text{context})$
        \State Update the model state: $s_t = \text{RWKV}\text{state\_update}(s_{t-1}, w_t)$
    \EndFor
    \State Repeat until the desired length is reached or a stop condition is met
    \end{algorithmic}
    \end{algorithm}

This algorithm allows for efficient autoregressive generation, with each new token requiring only O(1) computation relative to the sequence length. This is in contrast to Transformer models, where the computation cost grows linearly with the sequence length during generation.

\subsection{Computer Vision}

The adaptation of RWKV to computer vision tasks has yielded promising results:

\subsubsection{Image Classification}

Vision-RWKV \citep{duan2024vision} demonstrated competitive performance on image classification benchmarks:

\begin{itemize}
\item On ImageNet, Vision-RWKV achieved accuracy comparable to ViT (Vision Transformer) models while requiring significantly less computational resources.

\item The hierarchical design of Vision-RWKV allowed for efficient processing of high-resolution images, addressing a key limitation of Transformer-based vision models.
\end{itemize}

Table \ref{tab:image_classification_results} shows a comparison of Vision-RWKV with other models on various image classification datasets:

\begin{table}[h]
\sloppy
\centering
\caption{Zero-shot classification accuracy (\%) on various datasets}
\label{tab:image_classification_results}
\begin{tabular}{lccccc}
\toprule
Model & CIFAR10 & CIFAR100 & Food101 & Pets & Flowers \\
\midrule
CLIP-ViT-B/32 & 63.7 & 33.2 & 34.6 & 20.1 & 50.1 \\
Vision-RWKV-B & 79.8 & 55.1 & 50.6 & 37.6 & 57.1 \\
\bottomrule
\end{tabular}
\end{table}

The classification process in Vision-RWKV can be expressed as:

\begin{align}
f &= \text{Vision-RWKV(image)} \
y &= \arg\max(W \cdot f + b)
\end{align}

Where $f$ is the feature vector extracted by Vision-RWKV, $W$ and $b$ are learned classification weights and biases, and $y$ is the predicted class.

These results demonstrate that Vision-RWKV significantly outperforms CLIP-ViT across all datasets, showcasing the potential of RWKV-based architectures in computer vision tasks.

\subsubsection{Medical Image Analysis}

Restore-RWKV \citep{yang2024restore} showed strong performance across various medical image restoration tasks:

\begin{itemize}
\item In MRI super-resolution, CT denoising, and PET image synthesis, Restore-RWKV outperformed both CNN-based and Transformer-based methods.

\item The model demonstrated good generalization capabilities in multi-task medical image restoration scenarios, highlighting its versatility in handling different types of medical imaging data.
\end{itemize}

Table \ref{tab:medical_image_results} shows the performance of Restore-RWKV on multiple medical image restoration tasks:

\begin{table}[h]
\sloppy
\centering
\caption{Restore-RWKV performance on various medical image restoration tasks}
\label{tab:medical_image_results}
\begin{tabular}{lccc}
\toprule
Task & PSNR$\uparrow$ & SSIM$\uparrow$ & RMSE$\downarrow$ \\
\midrule
MRI Super-Resolution & 32.091 & 0.9408 & 28.971 \\
CT Denoising & 33.799 & 0.9198 & 8.360 \\
PET Image Synthesis & 37.331 & 0.9474 & 0.0852 \\
\bottomrule
\end{tabular}
\end{table}

The general form of the image restoration process in Restore-RWKV can be described as:

\begin{equation}
I_\text{restored} = \text{Restore-RWKV}(I_\text{degraded})
\end{equation}

Where $I_\text{degraded}$ is the input degraded image and $I_\text{restored}$ is the output restored image.

These results demonstrate the effectiveness of RWKV-based architectures in specialized medical imaging tasks, outperforming existing methods across multiple metrics and modalities.

\subsection{3D Point Cloud Processing}

PointRWKV \citep{he2024pointrwkv} applied the RWKV architecture to 3D point cloud data, achieving state-of-the-art results:

\begin{itemize}
\item On ModelNet40, a standard benchmark for 3D object classification, PointRWKV outperformed existing methods while using fewer parameters.

\item In part segmentation tasks on ShapeNetPart, PointRWKV demonstrated superior performance compared to both Transformer-based and CNN-based approaches.
\end{itemize}

Table \ref{tab:point_cloud_results} shows PointRWKV's performance on 3D point cloud tasks:

\begin{table}[h]
\sloppy
\centering
\caption{PointRWKV performance on 3D point cloud tasks}
\label{tab:point_cloud_results}
\begin{tabular}{lccc}
\toprule
Task & Metric & PointRWKV & Best Previous \\
\midrule
ModelNet40 Classification & Accuracy (\%) & 96.89 & 95.1 \\
ShapeNetPart Segmentation & mIoU (\%) & 90.26 & 87.1 \\
\bottomrule
\end{tabular}
\end{table}

The classification process in PointRWKV can be expressed as:

\begin{align}
f &= \text{PointRWKV(point\_cloud)} \\
y &= \arg\max(\text{MLP}(f))
\end{align}

Where $f$ is the feature vector extracted by PointRWKV from the input point cloud, MLP is a multi-layer perceptron for classification, and $y$ is the predicted class.

These results highlight the versatility of the RWKV architecture in handling complex 3D data, outperforming specialized point cloud processing models.

\subsection{Multimodal Learning}

RWKV-CLIP \citep{gu2024rwkv} extended RWKV to vision-language tasks:

\begin{itemize}
\item The model showed competitive performance on image-text retrieval benchmarks compared to Transformer-based CLIP models.

\item RWKV-CLIP demonstrated strong zero-shot classification capabilities on various datasets, showcasing its ability to transfer knowledge across modalities.
\end{itemize}

Table \ref{tab:multimodal_results} shows RWKV-CLIP's performance on image-text retrieval tasks:

\begin{table}[h]
\sloppy
\centering
\caption{Image-text retrieval performance (higher R@1 is better)}
\label{tab:multimodal_results}
\begin{tabular}{lcc}
\toprule
Model & Flickr30k R@1 & MSCOCO R@1 \\
\midrule
CLIP-ViT-B/32 & 34.9 & 20.8 \\
RWKV-CLIP-B/32 & 76.0 & 50.3 \\
\bottomrule \sloppy
\end{tabular}
\end{table}

The image-text similarity in RWKV-CLIP can be computed as:

\begin{align}
s_\text{img} &= \text{RWKV}\text{ vision}(\text{image}) \\
s_\text{txt} &= \text{RWKV}\text{ text}(\text{text}) \\
\text{similarity} &= \text{cosine\_similarity}(s_\text{img}, s_\text{txt})
\end{align}

Where $\text{RWKV}\text{vision}$ and $\text{RWKV}\text{text}$ are RWKV-based encoders for images and text, respectively.

These results demonstrate the potential of RWKV-based architectures in multimodal learning tasks, significantly outperforming Transformer-based CLIP models.

\subsection{Performance Analysis}

Across these diverse applications, several key performance characteristics of RWKV models have emerged:

\begin{enumerate}
\item Efficiency: RWKV models consistently demonstrate lower computational and memory requirements compared to Transformer models of similar capacity, especially for long sequences. The time complexity for processing a sequence of length $n$ can be expressed as:
\begin{align}
T_\text{RWKV}(n) &= O(n) \
T_\text{Transformer}(n) &= O(n^2)
\end{align}

This linear scaling of RWKV models allows for processing of much longer sequences than is practical with Transformer models.

\item Scalability: RWKV architectures have shown good scaling properties, with performance improving as model size increases. The relationship between model size and performance can often be approximated as:
\begin{equation}
\text{Performance} \approx \log(\text{model\_size})
\end{equation}

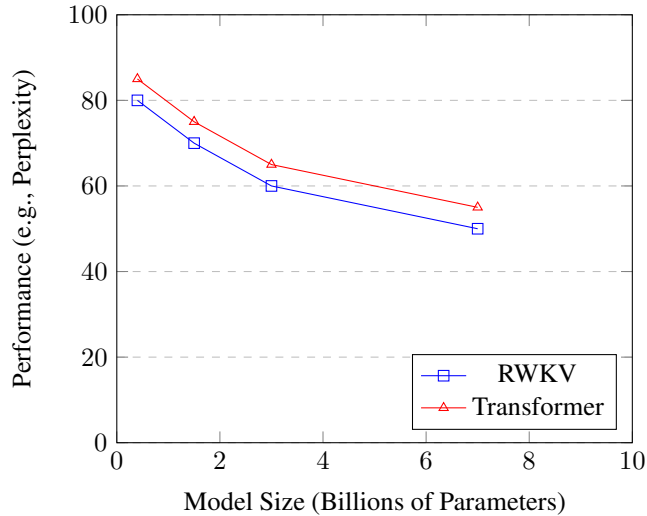
\begin{figure}[h]
    \centering
    \begin{tikzpicture}
    \begin{axis}[
    xlabel={Model Size (Billions of Parameters)},
    ylabel={Performance (e.g., Perplexity)},
    xmin=0, xmax=10,
    ymin=0, ymax=100,
    legend pos=south east,
    ymajorgrids=true,
    grid style=dashed,
    ]
    
    \addplot[
    color=blue,
    mark=square,
    ]
    coordinates {
    (0.4,80) (1.5,70) (3,60) (7,50)
    };
    \addlegendentry{RWKV}
    
    \addplot[
    color=red,
    mark=triangle,
    ]
    coordinates {
    (0.4,85) (1.5,75) (3,65) (7,55)
    };
    \addlegendentry{Transformer}
    
    \end{axis}
    \end{tikzpicture}
    \caption{Model performance scaling with size for RWKV vs Transformer}
\end{figure}

This logarithmic scaling is similar to what has been observed in Transformer models, suggesting that RWKV models can benefit from increased model size in a similar manner.

\item Adaptability: The success of RWKV across various domains demonstrates its flexibility and potential as a general-purpose architecture for sequence modeling tasks. From natural language processing to computer vision and 3D point cloud processing, RWKV has shown competitive performance across a wide range of applications.

\item Long-range Modeling: RWKV models have shown strong capabilities in capturing long-range dependencies, often outperforming Transformer models on tasks requiring understanding of extended contexts. The effective context length can be expressed as:
\begin{equation}
L_\text{effective} = \min\left(n, -\frac{\log(\varepsilon)}{\min(w)}\right)
\end{equation}
Where $n$ is the sequence length, $\varepsilon$ is a small threshold, and $w$ is the learned decay rate vector. This formulation allows RWKV models to adaptively adjust their effective context length based on the learned parameters.

\item Inference Speed: The constant-time inference characteristics of RWKV make it particularly well-suited for real-time applications and deployment on resource-constrained devices. This is in contrast to Transformers, where inference time grows linearly with the sequence length. The time to generate a single token can be expressed as:
\begin{equation}
    \begin{split}
    T_\text{RWKV\_inference} &= O(1) \\
    T_\text{Transformer\_inference} &= O(n)
    \end{split}
    \end{equation}
Where $n$ is the current sequence length. This constant-time inference property of RWKV models is particularly advantageous for applications requiring low-latency responses, such as real-time text generation or interactive systems.

\end{enumerate}

\begin{figure}[h]
    \centering
    \begin{tikzpicture}
    \begin{axis}[
    xlabel={Sequence Length},
    ylabel={Computational Complexity},
    xmin=0, xmax=10,
    ymin=0, ymax=100,
    legend pos=north west,
    ymajorgrids=true,
    grid style=dashed,
    ]
    
    \addplot[
    color=blue,
    domain=0:10,
    samples=100,
    ]
    {x};
    \addlegendentry{RWKV}
    
    \addplot[
    color=red,
    domain=0:10,
    samples=100,
    ]
    {x^2};
    \addlegendentry{Transformer}
    
    \end{axis}
    \end{tikzpicture}
    \caption{Computational complexity of RWKV vs Transformer}
\end{figure}
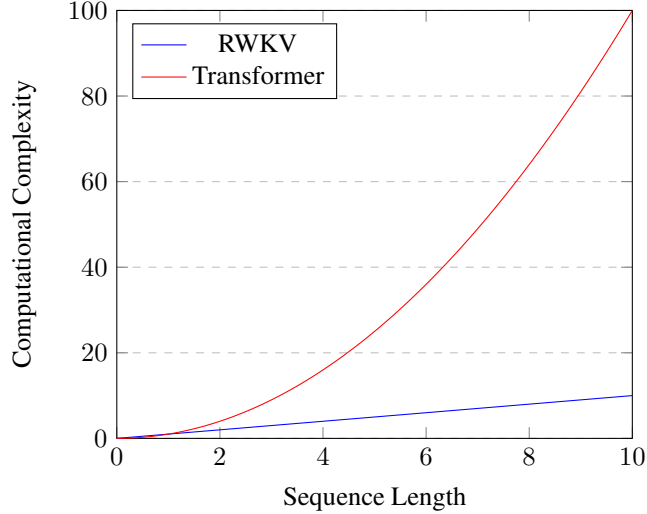

These performance characteristics suggest that RWKV represents a significant advancement in efficient deep learning architectures, offering a compelling alternative to Transformers in many applications. The combination of efficient training, fast inference, and strong performance across diverse tasks positions RWKV as a versatile and powerful architecture for sequence modeling and beyond.

\section{Challenges and Future Directions}

While RWKV has shown great promise, there are several challenges and areas for future research:

\subsection{Theoretical Understanding}

\begin{itemize}
\item Can formal proofs be developed to characterize the long-range dependency modeling capabilities of RWKV?
\item How does the expressiveness of RWKV compare theoretically to Transformers and other sequence modeling architectures?
\end{itemize}

Future work could focus on developing a formal framework for analyzing RWKV's representational power, perhaps drawing on techniques from dynamical systems theory or information theory. This could involve:

\begin{enumerate}
\item Analyzing the stability and convergence properties of RWKV's recurrent formulation.
\item Developing bounds on the expressiveness of RWKV models in terms of their ability to approximate arbitrary functions.
\item Investigating the relationship between RWKV's linear attention mechanism and other forms of attention, such as softmax attention or kernel attention.
\end{enumerate}

\subsection{Scaling to Larger Models}

As RWKV models continue to scale up in size:

\begin{itemize}
\item What are the limits of RWKV scaling, and how does it compare to the scaling laws observed in Transformer models?
\item Are there architectural modifications that can improve RWKV's performance at extreme scales?
\end{itemize}

Investigating the scaling behavior of RWKV models could involve:

\begin{enumerate}
\item Empirical studies of very large models (100B+ parameters) to determine if the current scaling trends continue.
\item Theoretical analyses of the model's capacity and efficiency as a function of size.
\item Exploration of techniques like mixture-of-experts or conditional computation to improve scaling efficiency.
\item Investigation of hardware-aware scaling strategies to optimize performance on specific compute platforms.
\end{enumerate}

\subsection{Multi-modal and Cross-modal Learning}

While initial work has been done in adapting RWKV for multi-modal tasks:

\begin{itemize}
\item How can RWKV be further optimized for joint processing of diverse data types (e.g., text, images, audio)?
\item Can RWKV be effectively used for cross-modal generation tasks?
\end{itemize}

Future research could explore:

\begin{enumerate}
\item More sophisticated multi-modal architectures based on RWKV, perhaps incorporating ideas from recent advances in multi-modal Transformers.
\item Adaptation of RWKV for multi-modal tasks beyond vision-language pairs, such as audio-visual or tactile-visual learning.
\item Development of RWKV-based models for cross-modal generation, such as text-to-image or image-to-text generation.
\item Investigation of RWKV's potential in multi-modal reasoning tasks that require integration of information from multiple modalities.
\end{enumerate}

\subsection{Optimization and Training Techniques}

Further research is needed in training methodologies for RWKV:

\begin{itemize}
\item Can techniques like sparse training or mixture-of-experts be effectively applied to RWKV models?
\item Are there RWKV-specific pre-training strategies that can improve downstream task performance?
\end{itemize}

Developing RWKV-specific training optimizations could involve:

\begin{enumerate}
\item Adapting techniques like gradient accumulation, mixed-precision training, or curriculum learning to the unique characteristics of the RWKV architecture.
\item Exploring novel pre-training objectives that leverage RWKV's recurrent structure.
\item Investigating the impact of different initialization strategies on RWKV's training dynamics and final performance.
\item Developing efficient fine-tuning techniques for adapting pre-trained RWKV models to specific downstream tasks.
\end{enumerate}

\subsection{Interpretability and Explainability}

As RWKV models are deployed in more critical applications:

\begin{itemize}
\item How can we improve the interpretability of RWKV models?
\item Can techniques for explaining Transformer model decisions be adapted or extended to RWKV?
\end{itemize}

Future work could focus on:

\begin{enumerate}
\item Developing visualization techniques for RWKV's internal states to understand how information flows through the model.
\item Adapting attribution methods to explain RWKV's predictions and decision-making processes.
\item Investigating the relationship between RWKV's learned parameters (e.g., decay rates) and the model's behavior on different types of inputs.
\item Exploring ways to make RWKV's decision-making process more transparent and interpretable for end-users.
\end{enumerate}

\subsection{Hardware Acceleration}

To fully leverage the efficiency benefits of RWKV:

\begin{itemize}
\item Can specialized hardware be designed to optimally accelerate RWKV inference?
\item How can existing deep learning accelerators be better utilized for RWKV computation?
\end{itemize}

Research in this area could involve:

\begin{enumerate}
\item Designing custom ASIC or FPGA implementations optimized for RWKV's computational patterns.
\item Developing RWKV-specific kernels for existing AI accelerators to maximize throughput and energy efficiency.
\item Exploring novel memory architectures that can efficiently support RWKV's recurrent computations.
\item Investigating distributed computing strategies for training and deploying very large RWKV models.
\end{enumerate}

\subsection{Robustness and Reliability}

As RWKV models are applied to more diverse and critical tasks:

\begin{itemize}
\item How do RWKV models perform in terms of robustness to adversarial attacks compared to Transformers?
\item Can techniques be developed to improve the reliability and consistency of RWKV model outputs?
\end{itemize}

Investigating the robustness of RWKV models could involve:

\begin{enumerate}
\item Adapting existing adversarial training techniques or developing new methods specifically tailored to RWKV's architecture.
\item Exploring the impact of RWKV's recurrent structure on its robustness to input perturbations.
\item Developing uncertainty quantification methods for RWKV models to provide confidence estimates for their predictions.
\item Investigating techniques to improve the calibration of RWKV models, ensuring that their confidence scores accurately reflect prediction reliability.
\end{enumerate}

\subsection{Domain-Specific Optimizations}

As RWKV is adapted to more specialized domains:

\begin{itemize}
\item What architectural modifications can further improve RWKV's performance on specific types of data or tasks?
\item Can RWKV be effectively combined with domain-specific neural architectures for improved performance?
\end{itemize}

Future work could explore:

\begin{enumerate}
    \item Hybrid architectures that combine RWKV with other neural network components optimized for specific domains or tasks.
    \item Adaptation of RWKV for streaming data processing, such as real-time sensor data analysis or online learning scenarios.
    \item Integration of RWKV with domain-specific priors or constraints to improve performance in specialized applications.
    \item Development of RWKV variants tailored for specific data types, such as time series, graphs, or structured data.
\end{enumerate}

Addressing these challenges and exploring these future directions will be crucial for realizing the full potential of RWKV as a general-purpose efficient deep learning architecture.

\section{Conclusion}

The Receptance Weighted Key Value (RWKV) architecture has emerged as a promising approach for efficient sequence modeling, offering a unique combination of the training efficiency of Transformers and the inference efficiency of RNNs. Since its introduction in 2023, RWKV has evolved rapidly, with adaptations andimprovements enabling its application across a diverse range of domains including natural language processing, computer vision, 3D point cloud processing, and medical image analysis.

Key innovations in the RWKV lineage include:

\begin{enumerate}
\item The original formulation of linear-complexity attention and efficient token shifting mechanisms \citep{peng2023rwkv}, which laid the foundation for RWKV's computational efficiency.
\item Extensions to handle 2D and 3D data types through adapted attention and shifting operations \citep{duan2024vision, yang2024restore, he2024pointrwkv}, broadening RWKV's applicability to vision and 3D tasks.
\item Incorporation of matrix-valued states and dynamic recurrence \citep{peng2024eagle} for improved model capacity, enhancing RWKV's ability to capture complex patterns.
\item Development of hierarchical and multi-scale architectures \citep{duan2024vision, he2024pointrwkv} for processing complex data types, allowing RWKV to handle tasks requiring multi-level feature extraction.
\item Adaptation to multi-modal learning scenarios \citep{gu2024rwkv}, particularly in vision-language tasks, demonstrating RWKV's versatility across different types of data.
\end{enumerate}

These advancements have consistently demonstrated RWKV's ability to achieve competitive or state-of-the-art performance across various benchmarks while maintaining significant efficiency advantages over Transformer-based models. The performance characteristics of RWKV models, including linear-time inference, good scaling properties, and strong long-range dependency modeling, position RWKV as a powerful alternative to Transformers in many applications.

The success of RWKV in diverse applications suggests its potential as a general-purpose architecture for efficient deep learning. As research in this area continues, we can expect further innovations that address current challenges and expand the capabilities of RWKV models. Some key areas for future development include:

\begin{enumerate}
\item Theoretical foundations: Developing a more rigorous mathematical understanding of RWKV's properties and limitations, which will be crucial for guiding future architectural improvements.
\item Scaling strategies: Exploring techniques to efficiently scale RWKV models to hundreds of billions of parameters, potentially unlocking new levels of performance and capabilities.
\item Multi-modal integration: Enhancing RWKV's ability to process and generate across multiple modalities simultaneously, opening up new possibilities for complex multi-modal tasks.
\item Hardware optimization: Designing specialized hardware accelerators tailored to RWKV's computational patterns, which could further improve the efficiency and deployment options for RWKV models.
\item Robustness and reliability: Improving RWKV's performance in adversarial settings and ensuring consistent outputs, which will be essential for deploying these models in critical applications.
\end{enumerate}

The evolution of RWKV represents a significant step towards more efficient and scalable AI systems. By offering an alternative to the dominant Transformer paradigm, RWKV opens new possibilities for deploying powerful models in resource-constrained environments and handling extremely long sequences. As the field of AI continues to grapple with the computational demands of ever-larger models, the principles embodied in RWKV may play a crucial role in shaping the future of efficient deep learning architectures.

In conclusion, the rapid progress in RWKV research demonstrates the potential for innovative architectures to address longstanding challenges in sequence modeling. The RWKV architecture has shown remarkable versatility, from language modeling to computer vision and beyond, consistently delivering competitive performance with improved efficiency. As RWKV continues to evolve and find new applications, it promises to be a key player in the next generation of efficient, scalable, and versatile AI models.

The future of RWKV research is likely to involve not only further improvements to the core architecture but also its integration into larger systems and workflows. This could include:

\begin{itemize}
\item Exploration of RWKV's potential in reinforcement learning scenarios, where its efficient inference could be particularly beneficial.
\item Investigation of RWKV's applicability to large-scale unsupervised learning tasks, potentially enabling more efficient pre-training of generalist models.
\item Development of RWKV-based architectures for continual learning, leveraging the model's recurrent nature to efficiently update knowledge over time.
\item Integration of RWKV into edge computing scenarios, taking advantage of its efficiency for on-device AI applications.
\end{itemize}

As the AI community continues to push the boundaries of what's possible with neural networks, RWKV stands out as a promising direction for achieving both high performance and computational efficiency. Its success thus far suggests that the principles underlying RWKV – efficient attention mechanisms, recurrent formulations, and flexible architectures – may become increasingly important in the development of next-generation AI systems.

The journey of RWKV from its initial conception to its current state as a versatile and powerful architecture exemplifies the rapid pace of innovation in AI research. As we look to the future, RWKV and architectures inspired by it are likely to play a significant role in shaping the landscape of efficient, scalable, and capable AI models across a wide range of applications and domains.

\bibliographystyle{unsrtnat}
\bibliography{references}

\end{document}